\documentclass[preprint]{elsarticle}

\usepackage{lineno}
\usepackage{graphicx}
\usepackage{dcolumn}
\usepackage{amsmath}
\usepackage{amssymb}
\usepackage{bm}
\usepackage{caption}
\usepackage{subcaption}
\usepackage{xcolor}

\journal{Nucl. Instr. and Meth. in Physics Res. Section A}
\begin{document}

\title{Unsupervised anomaly detection in MeV ultrafast electron diffraction}


\author[UStrath]{Mariana A. Fazio\corref{correspondence}}
\affiliation[UStrath]{Department of Electronic and Electrical Engineering, University of Strathclyde, Glasgow, UK}
\cortext[correspondence]{Corresponding author}
\ead{marianaafazio@gmail.com}

\author[UNM]{Manel Mart\'inez-Ram\'on}
\affiliation[UNM]{Department of Electrical and Computer Engineering, University of New Mexico, New Mexico, USA}

\author[UNM]{Salvador Sosa G\"{u}itron}%

\author[Brook]{Marcus Babzien}

\author[Brook]{Mikhail Fedurin}
\affiliation[Brook]{Brookhaven National Laboratory, Upton, New York, USA}

\author[Brook]{Junjie Li}

\author[Brook]{Mark Palmer}

\author[EA]{Sandra S. Biedro\'n}
\affiliation[EA]{Element Aero, San Leandro, California, USA}

\date{\today}

\begin{abstract}
MeV ultrafast electron diffraction (MUED) is a pump-probe technique used to study the dynamic structural evolution of materials. An ultrashort laser pulse triggers structural changes, which are then probed by an ultrashort relativistic electron beam. To overcome low signal-to-noise ratios, diffraction patterns are averaged over thousands of shots. However, shot-to-shot instabilities in the electron beam can distort individual patterns, introducing uncertainty. Improving MUED accuracy requires detecting and removing these anomalous patterns from large datasets. In this work, we developed a fully unsupervised methodology for the detection of anomalous diffraction patterns. Using a convolutional autoencoder, we calculate the reconstruction mean squared error of the diffraction patterns. Based on the statistical analysis of this error, we provide the user an estimation of the probability that the pattern is normal, which also allows a posterior visual inspection of the images that are difficult to classify. This method has been trained with only 100 diffraction patterns and tested on 1521 patterns, resulting in a false positive rate between 0.2\% and 0.4\%, with a training time of 10 seconds per image and a test time of about 1 second per image. The proposed methodology can also be applied to other diffraction techniques in which large datasets are collected that include faulty images due to instrumental instabilities.

\end{abstract}

\maketitle

\section{\label{sec:intro}Introduction}


MeV ultrafast electron diffraction (MUED) is a pump-probe characterization technique for studying ultrafast processes in materials {\  \cite{wang2003femto,zhu2015femtosecond, weathersby2015mega, qi2020breaking}.} The use of relativistic beams leads to decreased space-charge effects with respect to typical ultrafast electron diffraction experiments employing energies in the keV range. Compared to other ultrafast probes such as X-ray free electron lasers, MUED has a higher scattering cross section with material samples and allows access to higher-order reflections in the diffraction patterns due to the short electron wavelengths. However, this is a relatively young technology and several factors contribute to making it challenging to utilize, such as beam instabilities which can lower the effective spatial and temporal resolution.

In a typical MUED experiment, diffraction patterns are averaged over many shots to increase the signal-to-noise ratio. However, shot-to-shot instabilities in the electron beam can distort individual patterns leading to lower resolution in the averaged diffraction patterns. {\ These faulty or anomalous diffraction patterns, arising from drifts in the electron beam, are averaged into the dataset; in our acquisition chain, images are streamed and handled at an effective throughput of ~25 Mb/s (data rate) at 5 Hz repetition.} The resulting loss in resolution can obscure subtle structural changes in the material, especially in long experiments where significant drifts are expected to arise. Detection and elimination of these anomalous diffraction patterns from the large datasets generated by MUED can pave the way into increasing the resolution of the technique to explore ultrafast material processes with exceptional accuracy.

In the past years, machine learning (ML) approaches to materials and characterization techniques have provided a new path towards unlocking new physics by improving existing probes and increasing the user's ability to interpret data. In particular, ML methods can be employed to control characterization probes in near-real time, acting as virtual diagnostics, or ML can be deployed to extract features and lead to enhanced data analysis. ML up to now has yet to be applied to the MUED systems, where it can certainly enable advances that can further our understanding of ultrafast material processes in a variety of systems.

This study focuses on the construction of an unsupervised anomaly detection methodology to detect faulty images in MUED. We believe that unsupervised techniques are the best choice for our purposes because the data used to train the detector does not need to be manually labeled, and instead, the machine is intended to detect the anomalies in the dataset, which liberates the user from tedious, time-consuming initial image examination. The structure must, additionally, provide the user with some measure of uncertainty in the detection, so the user can take decisions based on this measure. If most images have a low uncertainty in the decision, the user may decide to manually examine a few images not confidently detected by the algorithm or directly throw them away to eliminate the associated risk of using them. 

Non supervised methods can be roughly grouped in density-based, bagging, deep learning, or kernel methods. The residual methods can use some of the above-mentioned techniques.  Density-based methods are those that estimate a compact observation model and from it a likelihood of these observations can be inferred. Then, the anomalous nature of a sample can be judged from this measure. In \cite{goldstein2012histogram}, histograms are used for this purpose, and in \cite{breunig2000lof}, dissimilarity measures between samples are computed to determine whether a sample is an outlier or not. The k-nearest neighborhood strategy is used in  \cite{ramaswamy2000efficient, angiulli2002fast} to provide an approximate measure of likelihood. The bagging models use an ensemble of detectors constructed with bagging or data subsampling  \cite{lazarevic2005feature,aggarwal2015theoretical}. 

In the field of deep learning, we find methods that are also non-supervised and that are intended to process the data to reduce its dimensionality in what is called a bottleneck while preserving its distribution properties. These approaches are convenient when the dimension of the data is very high, as it is the case of images. The autoencoder model \cite{hinton2006reducing} is used for this purpose in \cite{sakurada2014anomaly, chen2017outlier}. Generative adversarial networks \cite{goodfellow2020generative} are applied for anomaly detection in \cite{liu2019generative}.

Deep learning is useful when the number of data is high, but when the number of samples is low, we may take into consideration the use of kernel methods \cite{shawe2004kernel}. See, for example, the applications for outlier detection in  \cite{hardin2004outlier} or \cite{shyu2003novel}.  The latter uses an autoencoder together with nonlinear dimensionality reduction with the help of kernel methods. 

One of the most prominent kernel method for anomaly detection is perhaps the OC-SVM \cite{scholkopf2001estimating}, that makes use of the maximum margin criterion \cite{vapnik2013nature}, which effectively imposes a balance between the training error and the expressive capacity of the machine, reducing the overfitting when the number of samples is low. The OC-SVM is, besides, a non-supervised methodology. 
The Support Vector Data Description (SVDD)  in \cite{tax2004support} is based in the same criterion.

Residual methods are based on the examination of the prediction error over the sample under test. If a given variable that is observable can be inferred from the sample, then the prediction error can be measured. If the sample is anomalous, it is expected that it will produce a high error or residual. The determination of whether the residual is high or not can be achieved by a model of the residual distribution (see, e.g. \cite{zamzam2019data}). 

{\ Our approach is based on the use of a lightweight convolutional autoencoder (CAE) \cite{mao2016image}  that, with low computational burden, is able to reconstruct the input images and denoise them conveniently. The reconstruction error is used as a residual to estimate the posterior probability of that the image is normal conditional to the observed error. The error is modeled in a probabilistic way using a convex combination of distributions that automatically capture the conditional distribution of the error under the hypothesis of normal or fault.  The method provides added value with respect to a standard statistical evaluation in terms of its capability to denoise the images through the use of the CAE, together with an ability to estimate the probability that an image is faulty. Then, the user can discard the images with a high probability and only manually inspect those that have a probability close to 0.5, which are, as it is seen in the experimental Section, a very small fraction of the analyzed images. }


\section{\label{sec:experimental}Experimental Methods}

\begin{figure}[!h]
	\centering
	\includegraphics[width=\linewidth]{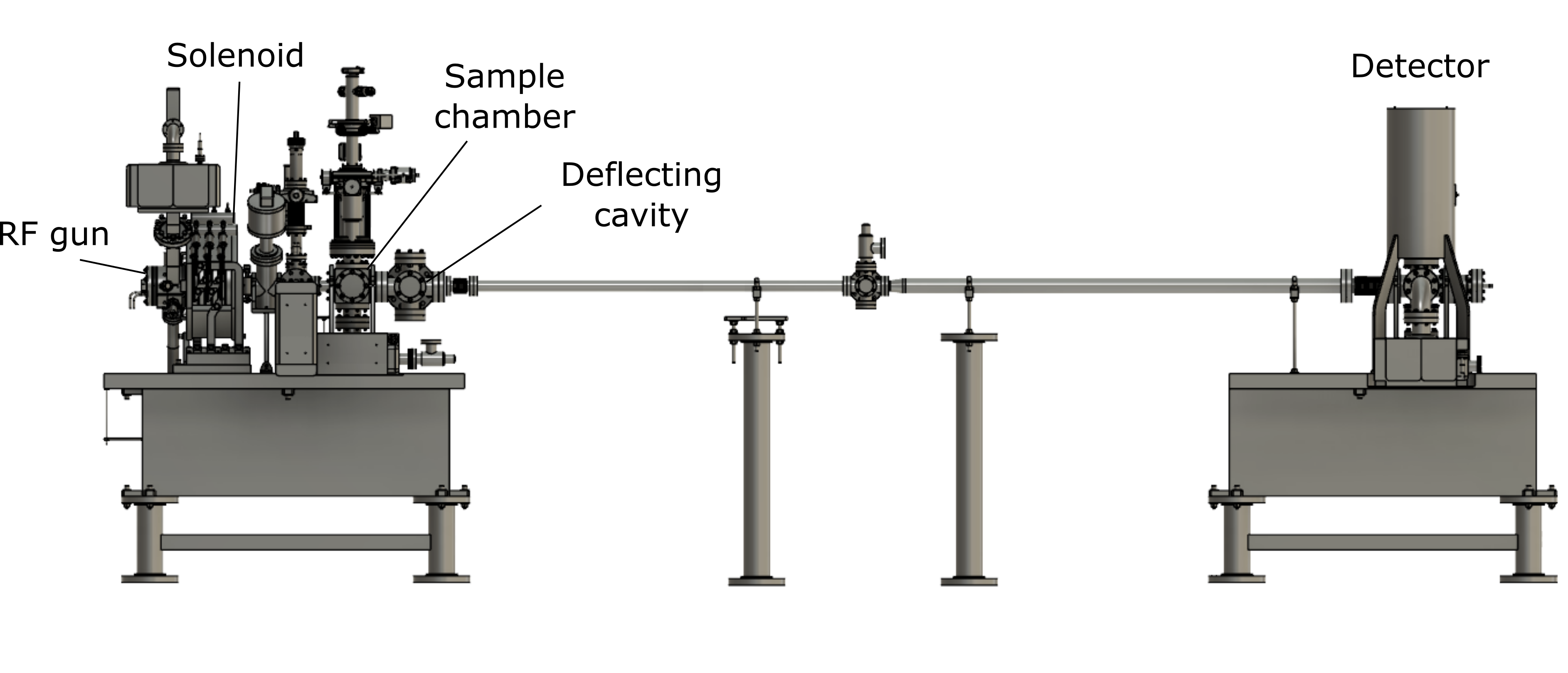}
	\caption{Schematic representation of the experimental setup for the MUED instrument located in the Accelerator Test Facility at Brookhaven National Laboratory.}
	\label{fig:mued}
\end{figure}

The MUED instrument is located in the Accelerator Test Facility at Brookhaven National Laboratory. A schematic representation of the experimental setup is presented in Fig. \ref{fig:mued}. The femtosecond electron beams are generated using a frequency-tripled Ti:Sapphire laser that illuminates a copper photocathode, generating a high brightness beam. The electrons are then accelerated and compressed in a 1.6-cell RF cavity achieving energies up to 5 MeV. Current parameters of the electron beam source optimized for stability are presented in Table \ref{tab:parameters}. The sample chamber is located downstream from the source with a motorized holder for up to nine samples with cryogenic cooling capabilities and a window to allow laser pumping of the material. Next to the chamber a RF deflecting cavity is located and 4 m downstream the detector system is placed to collect the diffraction patterns. The detector consists of a phosphor screen followed by a copper mirror (with a hole for non-diffracted electrons to pass) and a CCD Andor camera of 512 pixels x 512 pixels with a large aperture lens. {\ Taking into account that the numericalk aperturre of the lens was f/0.95 and the phosphor grain is less than 6 $\mu m$, the ratio between the Airy disk and the pixel size in this camera for green light is less than 1.}  Suitable material systems for MUED require careful preparation with typical lateral sizes of 100 - 300~$\mu$m and roughly < 100 nm thickness to assure electron transparency. Laser fluence is adjusted to avoid radiation-induced damage of the probed material.

\begin{table}[!hbt]
	\centering
	\caption{MUED source parameters for typical operation}
	\begin{tabular}{|l|c|}
		\hline
		\textbf{Beam energy} & 3 MeV\\
		\hline
		\textbf{Number of electrons per pulse} & 1.25 $\times$ 10$^6$\\
		\hline
		\textbf{Temporal resolution} & 180 fs \\
		\hline
		\textbf{Beam diameter} & 100 - 300 $\mu$m \\
		\hline
		\textbf{Repetition rate} & 5 - 48 Hz\\
		\hline
		\textbf{Number of electrons per sec per $\mu$m$^2$} & 88 - 880\\
		\hline
	\end{tabular}
	\label{tab:parameters}
\end{table}

\begin{figure}[h!]
    \begin{subfigure}{.475\linewidth}
        \centering
        \includegraphics[width=\linewidth]{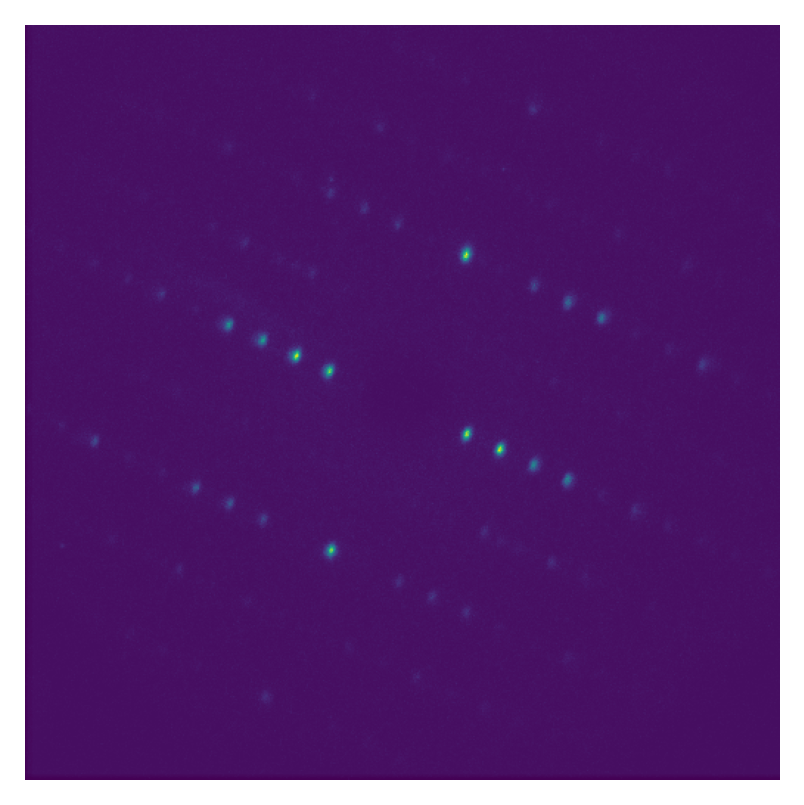}  
        \caption{}
        \label{fig:sub-first}
    \end{subfigure}
    \hfill
    \begin{subfigure}{.475\linewidth}
        \centering
        \includegraphics[width=\linewidth]{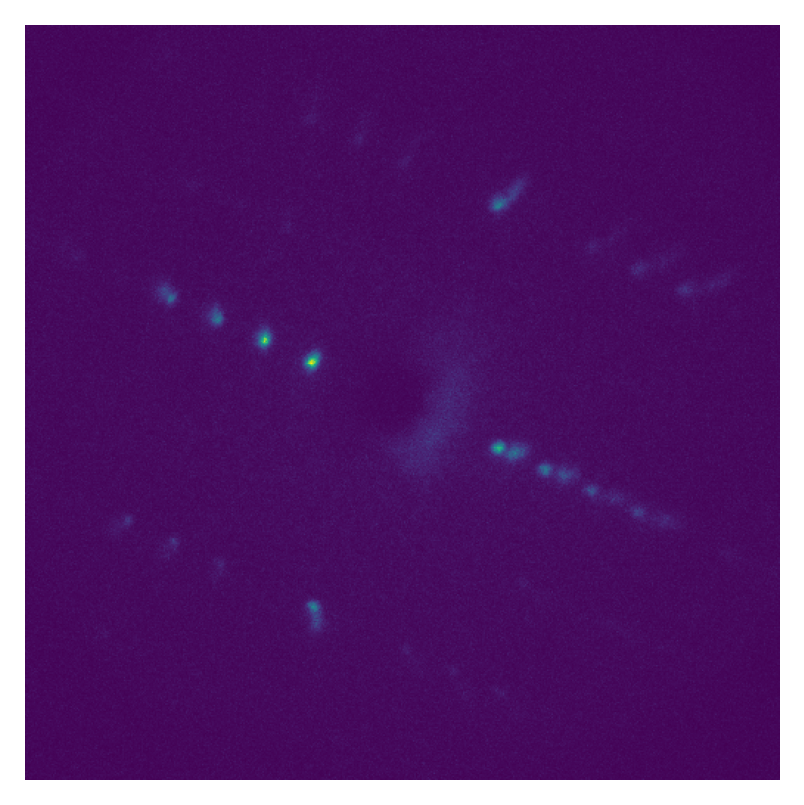}  
        \caption{}
        \label{fig:sub-second}
    \end{subfigure}
    \vskip\baselineskip
    \begin{subfigure}{.475\linewidth}
        \centering
        \includegraphics[width=\linewidth]{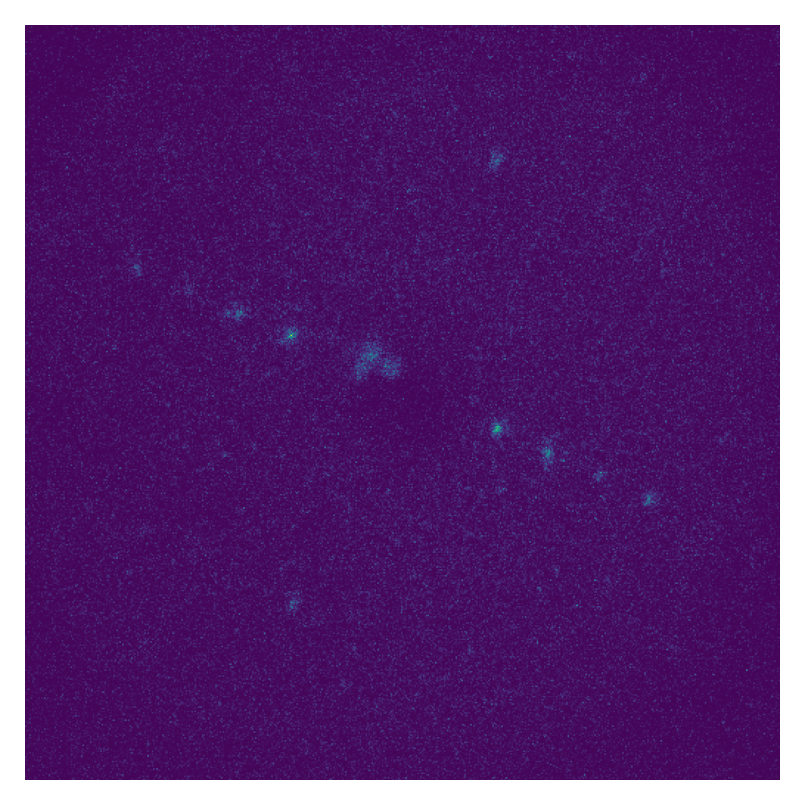}
        \caption{}
        \label{fig:sub-third}
    \end{subfigure}
    \hfill
    \begin{subfigure}{.475\linewidth}
        \centering
        \includegraphics[width=\linewidth]{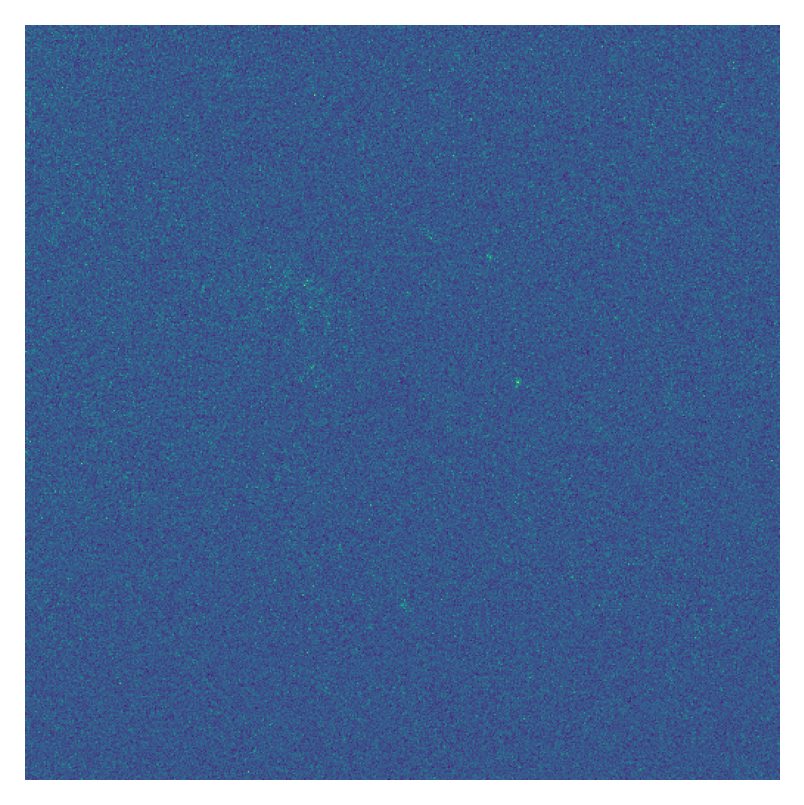}  
        \caption{}
        \label{fig:sub-fourth}
    \end{subfigure}
\caption{\label{fig:typical_pattern} Single shot diffraction patterns obtained for Ta$_2$NiSe$_5$: (a) typical diffraction pattern and (b) - (d) anomalous patterns.}
\end{figure}

In the present study, electron diffraction patterns were measured for a thin single crystal Ta$_2$NiSe$_5$ sample {\ \cite{baldini2023spontaneous, chen2025structural}}. No pump laser was employed and single shot diffraction patterns were captured. Figure \ref{fig:typical_pattern}(a) shows a typical diffraction pattern measured for Ta$_2$NiSe$_5$ consisting of a grayscale image of 512 x 512 pixels. Figures \ref{fig:typical_pattern}(b)-(d) present different types of anomalies due to instabilities in the electron beam. A typical dataset will contain between 70 - 90 single shot diffraction patterns, including such anomalies that will affect the accuracy of the experiment. For this work, several datasets were captured during multiple days of operation of the MUED experiment employing the Ta$_2$NiSe$_5$ sample.

\section{\label{sec:computational}Computational Methods}

\subsection{\label{sec:preprocessing}Data pre-processing procedure}

As can be observed in figure \ref{fig:typical_pattern}, a diffraction pattern for a single crystal material consists of mostly background superimposed with discrete Bragg peaks. In order for the model to learn effectively the features of interested of the diffraction patterns, input patterns should consist mostly of signal (Bragg peaks) with minimal background. Therefore, pre-processing is a key step for ensuring good performance of the models for reconstruction of the diffraction patterns.

\begin{figure}
    \centering
    \includegraphics[width=1\linewidth]{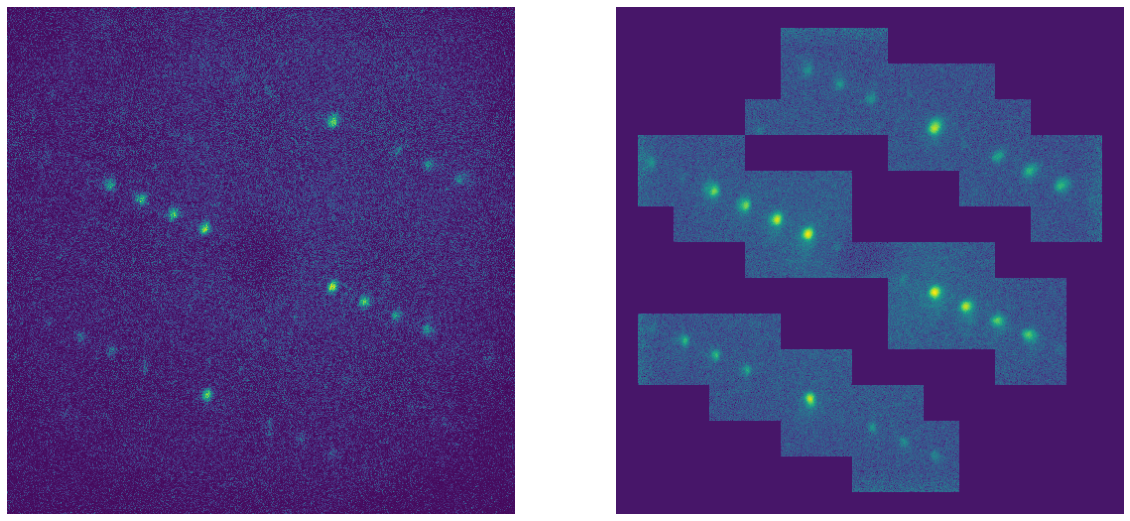}
    \caption{Original normal image and the corresponding tiled image after the preprocessing. The figure represents the logarithm of thre amplitude to enhance the details of the tiling.}
    \label{fig:original_tiled}
\end{figure}

Each input pattern was normalized and divided in overlapping tiles of 80 x 80 pixels. Tiles containing only background were filtered out of the dataset. Tiles can be defined as background if they contain white noise. To evaluate this, it is simpler to determine whether the image has a high probability of containing Bragg peaks by counting the number of connected pixels with an amplitude over a threshold computed as a function of the median of the {\ tile} amplitude. The used algorithm simply consists of detecting all pixels in the image with a value over $m$ times the median. If a pixel is copnnected to  $n$ or more other pixels over the same threshold, the image is selected. The chosen values were $m=3$ and $n=10$, which correctly selected all tiles containing Bragg peaks in a small sample.  An example of the result of image tiling can be seen in Fig. \ref{fig:original_tiled}, where the logarithm of the pixel amplitude is shown.

{\ The size of the images is set so that they are significantly larger than a Bragg spot, but small enough not to compromise the computational burden of the methodology during training.  As it will be detailed further, a choice of $80 \times 80$ pixels with the chosen architecture leads to a latent space of dimension 256 (see Table \ref{table:cae} below) after 3 convolutions. If we increase the dimension of the input, the dimension of the latent space increases quadratically, hence the computational burden.

To crossvalidate parameters $m$ and $n$, we conducted experiments with different integer values of $m$ from 1 to 4 and different values of $n$ equal to 3, 6, 10 and 20. In the 16 experiments, we visually detected the number of wrongly selected tiles among a set of 160 tiles. The combinations $m=3$
and $n= 6$ or 10 detected all tiles without error.

The tile discrimination was done during the training only, but during the test, all tiles were processed in order to be able to reconstruct full denoised images. }








\subsection{\label{sec:autoencoder}Convolutional autoencoder for reconstruction}

\begin{figure}
    \centering
    \includegraphics[width=0.8\linewidth]{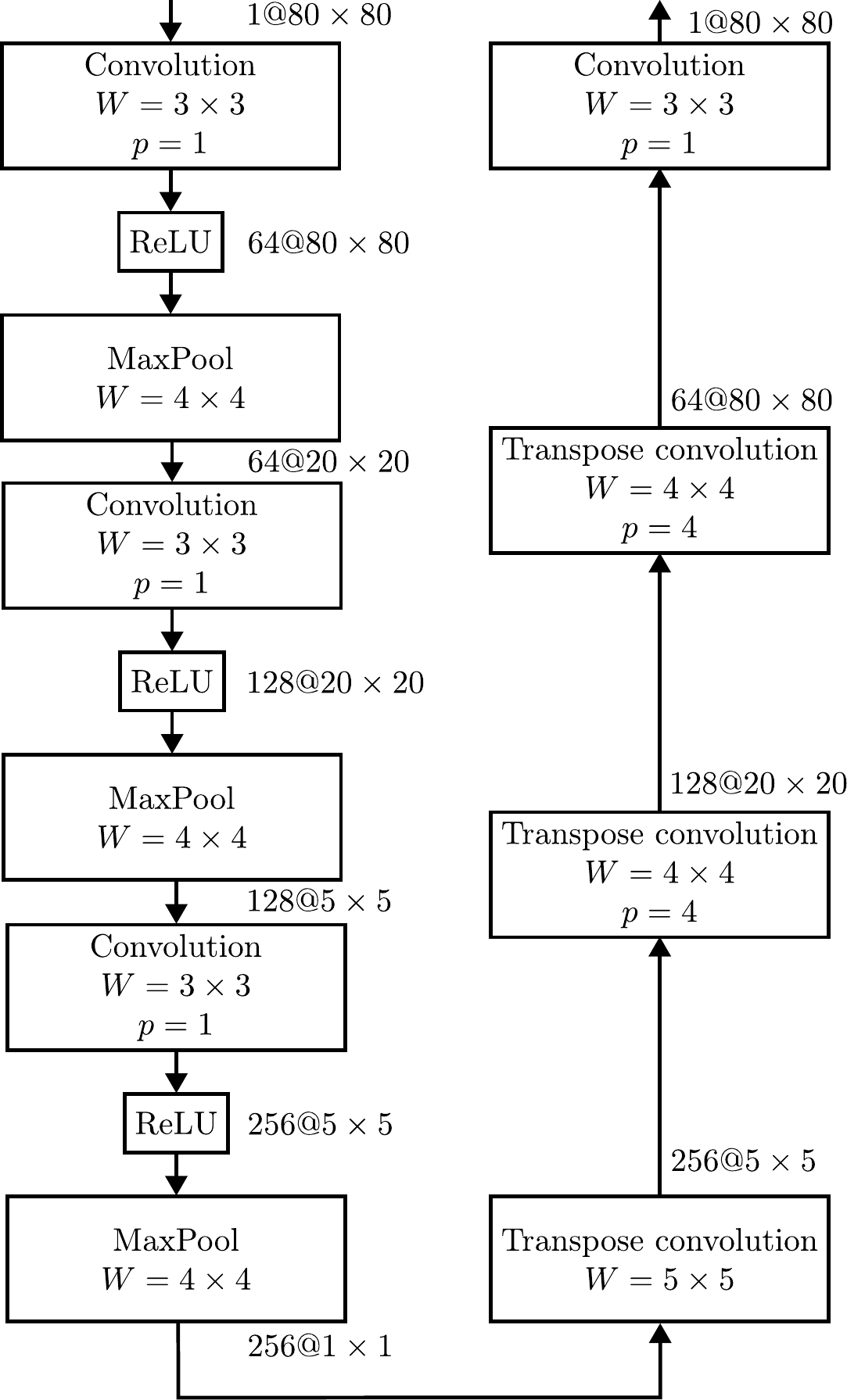}
    \caption{Structure of the used convolutional encoder. The input is a tile of an image, with dimensions $80\times 80$ for the present application, and the output is the reconstructed image contained in the tile.}
    \label{fig:CE}
\end{figure}

A convolutional autoencoder (CAE) has been used as the main element of the system. A schematic of the CAE model used for reconstruction of the diffraction patterns can be found in Fig. \ref{fig:CE} and the full structure and parameters are presented in Table \ref{table:cae}. The parameters of the CAE have been cross-validated with the use of a set of normal images. 

\begin{table}[h!]
\centering
\begin{tabular}{|l | c | c |} 
 \hline
 Layer & Output shape & Parameters \\ [0.5ex] 
 \hline\hline
 Input & (80, 80, 1) & 0 \\
 \hline
 Conv2D + ReLu & (80, 80, 64) & 640 \\
 MaxPool2D & (20, 20, 64) & 0 \\
 \hline
 Conv2D + ReLu & (20, 20, 128) & 73,856 \\
 MaxPool2D & (5, 5, 128) & 0 \\
 \hline
 Conv2D + ReLu & (5, 5, 256) & 295,168 \\
 MaxPool2D & (1, 1, 256) & 0 \\
 \hline
 ConvTranspose2D + ReLu & (5, 5, 256) & 1,638,400 \\
 \hline
 ConvTranspose2D + ReLu & (20, 20, 128) & 524,288 \\
 \hline
 ConvTranspose2D + ReLu & (80, 80, 64) & 131,072 \\
 \hline
 Conv2D  & (80, 80, 1) & 576 \\
 \hline
\end{tabular}
\caption{Structure and number of parameters of the convolutional autoenconder employed for reconstruction of the diffraction patterns.}
\label{table:cae}
\end{table}

The CAE features a total of 6 layers and the output of the enconder (the feature vector) contains  256 components. For the encoder, each layer consists of a 2D convolution operation \cite{gu2018recent}. For  convolution layer $\ell$ with $C_{\ell -1 }$ input channels and $C_\ell$ output channels,  the operation can be written as

\begin{equation}
    {\bf Z}^{\ell}_j = \sum_{i=1}^{C_{\ell-1}} {\bf W}^{\ell}_{i,j} * {\bf H}^{\ell-1}_i + {\bf B}_j^{\ell}, ~~1 \leq j \leq C_{\ell}
\end{equation}

This is, each one of the channels ${\bf H}_i^{\ell-1}$ of the previous layer is convolved with convolution kernel ${\bf W}^{\ell}_{i,j}$ and then added together. Then a matrix ${\bf B}_j^{\ell}$ of bias values is added to produce the output $ {\bf Z}^{\ell}_j$ of channel $j$ in later $\ell$. 

This image is then zero padded, this is, a number $p$ of rows containing zeros are added at the left and right and to top and bottom of the image.

This output is then passed through a nonlinear transformation with two steps,  a nonlinear activation and a MaxPooling  \cite{fukushima1982neocognitron, lecun1998gradient} to produce output channel $ {\bf H}^{\ell}_j$, that will be summarized further. 

The first convolution layer has 64 output channels and the size of each channel is $80\times 80$. The convolution kernel is $3\times 3$ and a zero padding of one pixel is applied to each edge of the resulting image. Therefore, the convolution output has a width and height of $80-3+1=78$. Due to the zero padding, the dimension is increased to $80$. After that, a MaxPooling is applied with a window of dimensions $4\times 4$. The MaxPooling consists of selecting the pixel with the maximum value inside a window of pixels, for all disjoint windows of the image. Therefore, the dimension of the resulting image is reduced by a factor 4 in each dimension. This results in a final dimension of $20\times 20$ pixels. After this, a Rectified Linear Unit (ReLU) activation \cite{nair2010rectified} is applied. The ReLU activation is the simplest nonlinear activation and it can be written as 
$$
\text{ReLU}(z) = max(0,z)
$$ 
this is, if the argument is negative, the output is zero, otherwise the output is equal to the input.  

Following the same computations, the output of the last convolutional layer contains images of a single pixel. The number of channels at the output is 256. This is the so called bottleneck of the encoder, and it is intended to contain all the information needed to reconstruct the image. 

For the decoder, each layer consists of a 2D transpose convolution operation \cite{dumoulin2016guide} with a rectified linear unit activation that simultaneously increases the dimensionalty and decreases the channels in a similar manner to the layers of the encoder. The operation can be written as

\begin{equation}
    {\bf Z}^{\ell}_j = \sum_{i=1}^{C_{\ell-1}} {\bf W}^{\ell}_{i,j} \otimes {\bf H}^{\ell-1}_i + {\bf B}_j^{\ell}, ~~1 \leq j \leq C_{\ell}
\end{equation}

In this operation, instead of a convolution, a Hadamard matrix product between  the kernel and the image is applied. The dimension of the resulting image is equal to the product of the dimension of the input image times the dimension of the kernel. The first transpose convolution has a kernel of dimensions $5\times 5$, and since the input has a single pixel, the resulting images have dimensions $5 \times 5$. The dimensionality is further increased with two more transpose convolutions up to $\times 80$ and 64 channels. In order to combine all the channels, a last convolution with a kernel of dimensions $4\times 4$ and zero padding with $p=1$ results in a single channel of dimension $80 \times 80$.  The output layer is a 2D convolution operation with a linear activation.
{\ The choice of the parameters of the CAE has not been fully explored, due to the limitations of the computational resources at hand, but the performance of  the present architecture can be improved by such crossvalidation of parameters. The use of larger images implies a higher number of parameters, at least in the input and output layers, which translates into convolutions of a much higher computational burden. With the maxpool sizes of the present architecture, it would take more convolutional layers to arrive to a small bottleneck (128 or 256). Also, the convolutions need to capture local relationships between a Bragg peak in order to proceed to the reconstruction, and therefore, we did not consider it necessary to use larger images. }

The mean squared error ($MSE$) loss was used with an Adam optimizer \cite{kingma2014adam} and standard backpropagation. {\ The justification of the use of the MSE criterion is simply based on the assumption that the error is Gaussian. If the error likelihood is Gaussian and a Gaussian prior is established for the set of network parameters, then the posterior probability of these parameters given the training data can be written as $p({\bf e}|{\mathcal X})\propto \exp\left(-\gamma {\boldsymbol \Theta}\right)\prod_i \exp\left(\frac{-\sigma^{-2}}{2} \|{\bf X}_i-{\bf {\hat X}_i}\|^2 \right)$. The first term is the prior and the second models the likelihood, where ${\bf X}_i,~{\bf {\hat X}_i}$ represent the input and reconstructed images, respectively. The minimization of the mean square error plus a regularization term is equivalent to maximizing the log posterior of the parameters, often called Maximum a Posteriori (MAP) criterion. Indeed, taking logarithms in both sides and negating them leads to the expression $\sum_i \frac{\sigma^{-2}}{2} \|{\bf X}_i-{\bf {\hat X}_i}\|^2  + \gamma {\boldsymbol \Theta}$, which contains the sum of the squared errors plus a regularization term,  that is included in the Adam optimization algorithm. }

\subsection{Anomaly detection}

The detection of anomalies is simply performed by the computation of the MSE error across all pixels of a reconstructed image. Low errors will correspond to normal images, and higher errors correspond to anomalies. The rational behind this strategy can be that the number of anomalous images during the training is negligible compared to the number of normal images. Then, the CAE can reconstruct these images, but the number of anomalies is too low for the CAE to be able to reconstruct them, since they are drawn from a different distribution. This anomaly detection method is known as residual analysis detection \cite{hwang2009survey}.  

Another explanation for the behavior of the MSE error is that the CAE has enough expressive capacity to reconstruct normal images, but the anomalies introduce additional complexity in the images, which the CAE is not able to learn. So to speak, our model is purportedly \textit{underfit} for the faulty images. In this case, the CAE can be trained with a significant number of anomalies, which will not affect the detection performance. There is an extensive scholarship in generalization bounds in deep learning (see, e.g. \cite{cao2019generalization}) that treat the topic of overfitting and underfitting in neural networks. 

If a significant number of anomalies are present in a dataset, the CAE will present a high training error in these images. In preliminary experiments, we observed that the errors group in two separate clusters, and therefore the likelihood function of these errors can be modeled with a mixture of probability density functions $p(e|N)$ and  $p(e|A)$ \cite{murphy2012machine}, which are the likelihoods of the error under the hypothesis that the image is normal ($N$) or an anomaly ($A$). If the prior probability for these anomalies $p(A)=1-p(N)$,  then a posterior probability can be written as

\begin{equation}\label{eq:posterior}
    p(N|e)=\frac{p(e|N)p(N)}{p(e|A)p(A)+p(e|N)p(N)}
\end{equation}

It has been experimentally shown that the marginal likelihood of the MSE error 

\begin{equation}
p(e)=p(e|A)p(A)+p(e|N)p(N)
\end{equation}
can be modeled with two Gamma distributions or two Rice distributions as conditional likelihoods. If it is assumed that a significant number of anomalies is present in a dataset, but they are not previously detected, as it is the case of this work, both the parameters of the distributions and the values of priors $p(N)$ and $p(A)$ can be easily adjusted by the use of the ME algorithm or by using the limited memory Broyden–Fletcher–Goldfarb–Shanno with box constraints (L-BFGS-B) \cite{fletcher2000practical} algorithm. {\ The details of the strategy to asses the convergence of the algorithm are explained in Subsection 4.2 below.}  

{\ It must be noted that in dynamic experiments, we must avoid treating real pump-induced changes as anomalies. The cleanest deployment is to train (or calibrate the error distribution) on a representative subset of nominally good frames for each condition (unpumped and pumped) or on pre-pump baseline frames, and then apply the detector within each condition/time-delay block. The model’s output probability  $p(N|e)$ then identifies frames that deviate from the typical appearance for that condition.}

\begin{figure}
    \centering
    \includegraphics[width=0.6\linewidth]{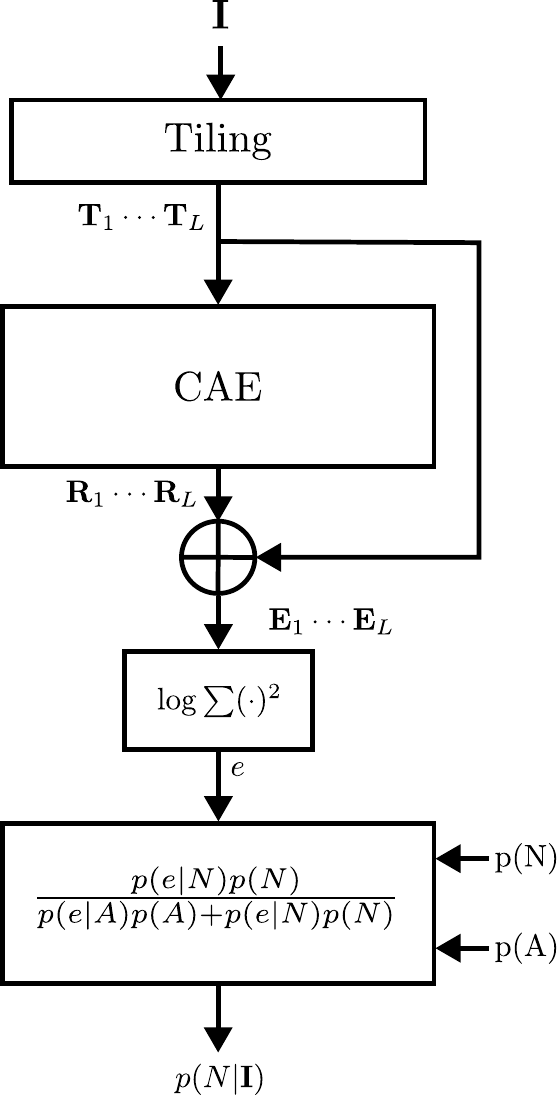}
    \caption{Structure of the detector. }
    \label{fig:detector}
\end{figure}
Once the posterior probability is estimated, the detector provides the user with an estimation of the probability that the sample is normal. That way, those samples that have a probability close to 0.5 can be visually analyzed by the user. 

The process is visualized in Fig. \ref{fig:detector}. The process starts with the introduction of an image, which is tiled. Each of the tiles ${\bf T}_i$ is reconstructed and its reconstruction error ${\bf E}_i$ is computed. Then, the logarithm of the mean squared error across all the pixels of the tiles is computed and with it, an estimation of the posterior $p(N|e)=p(N|{\bf I})$ is computed. In the system, it is assumed that both likelihoods and the priors are estimated. The details of the particular algorithm for this estimation are provided in the next section. 

Moreover, a threshold can be established by finding the value of the error $e$ that gives a certain probability, for example, $0.5$. The threshold can be used to classify the samples with a reasonable balance between probability of detection and probability of false alarm.  

\begin{figure*}
    \centering
    \includegraphics[width=0.9\linewidth]{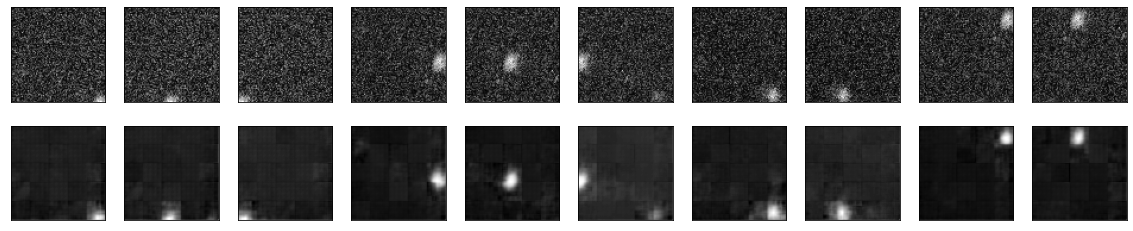}\\
    \includegraphics[width=0.9\linewidth]{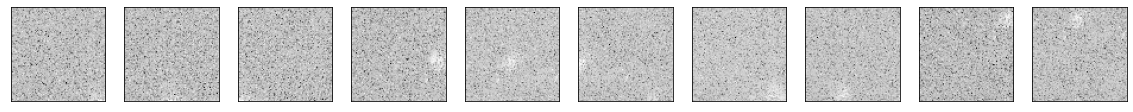}\\
    \includegraphics[width=0.9\linewidth]{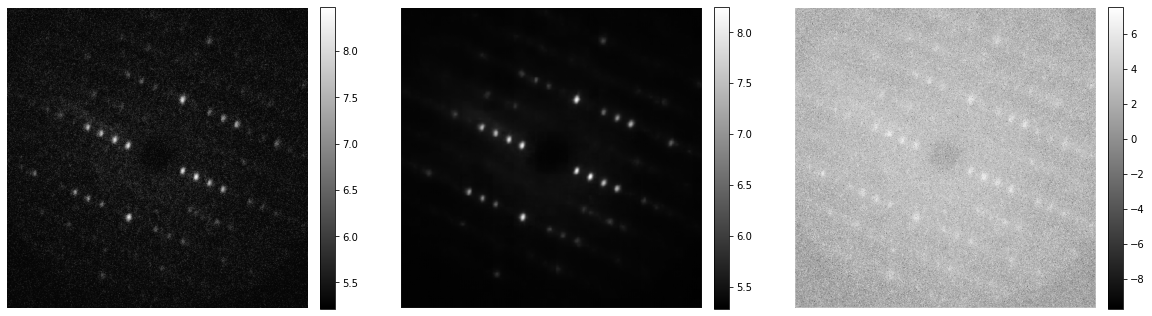}
    \caption{Comparison between 10 original and reconstructed test tiles of an image and their corresponding reconstruction error {\ and the entire reconstructed image. The graphs show the logarithm of the amplitudes and the squared error. The original and reconstructed features are shown in pixel coordinates}.}
    \label{fig:reconstruction_example}.
\end{figure*}
\subsection{\label{sec:alcf}Computational resources}
The experiments were run in the Center for Advanced Research Computiing (CARC) Supercomputer of the University of New Mexico. The used resources were a SGI AltixXE
Xeon X5550 at 2.67 GHz, with Intel Xeon Nehalem EP architecture. A total of 8 nodes with 8 cores each were used. The operating system was Linux CentOS 7. The programming language was Python 3 and the convolutional autoencoder was programmed using the Pytorch library. 
\section{Experiments}

\subsection{Training and test of the autoencoder}

{\ Training is performed using 100 training images (resulting in 5492 tiles) and 30 validation images (1520 tiles). The batch size is 32 tiles, which, given the current tiling parameters, corresponds to batches of 172 tiles. The training algorithm is the standard Adam with MMSE criterion. The training parameters are: $\mu=0.01$, and 10 training epochs. The total training time was of less than 2 minutes. The machine is agnostic about which images are faulty. The mean and standard deviation of the training images are computed, and then training and validation (and later test) images are normalized with these parameters. No further preprocessing is applied to the images. 

After each epoch, we evaluate performance on the 30 validation images. We observe that the validation error does not change significantly after 10 epochs; however, a more rigorous stopping strategy should rely on an explicit criterion as early stopping when the validation error reaches a minimum within a specified tolerance, or stopping upon detection of overfitting (i.e., when the validation error begins to increase).}

During the test, each image is segmented in tiles, {\ without discarding the ones not showing Bragg peaks}, and passed to the autoencoder. The mean squared error is computed across all pixels of all tiles of the image. 
 An example of the reconstruction for 10 arbitrary normal tiles of an image is shown in Fig. \ref{fig:reconstruction_example}, {\ where the original and reconstructed features are shown in pixel coordinates}.

The CAE was tested with the available 1521 images (including the training ones), of which 615 (40.4\%) of them were faulty images, and 906 were normal. The total number of analyzed tiles is about $10^5$. The total test time was of 24 minutes, corresponding roughly to 1 second per image. This includes the image loading from the hard drive, which took most of the processing time.  



\subsection{Statistical analysis of the reconstruction error}

\begin{figure}
    \centering
    \includegraphics[width=\linewidth]{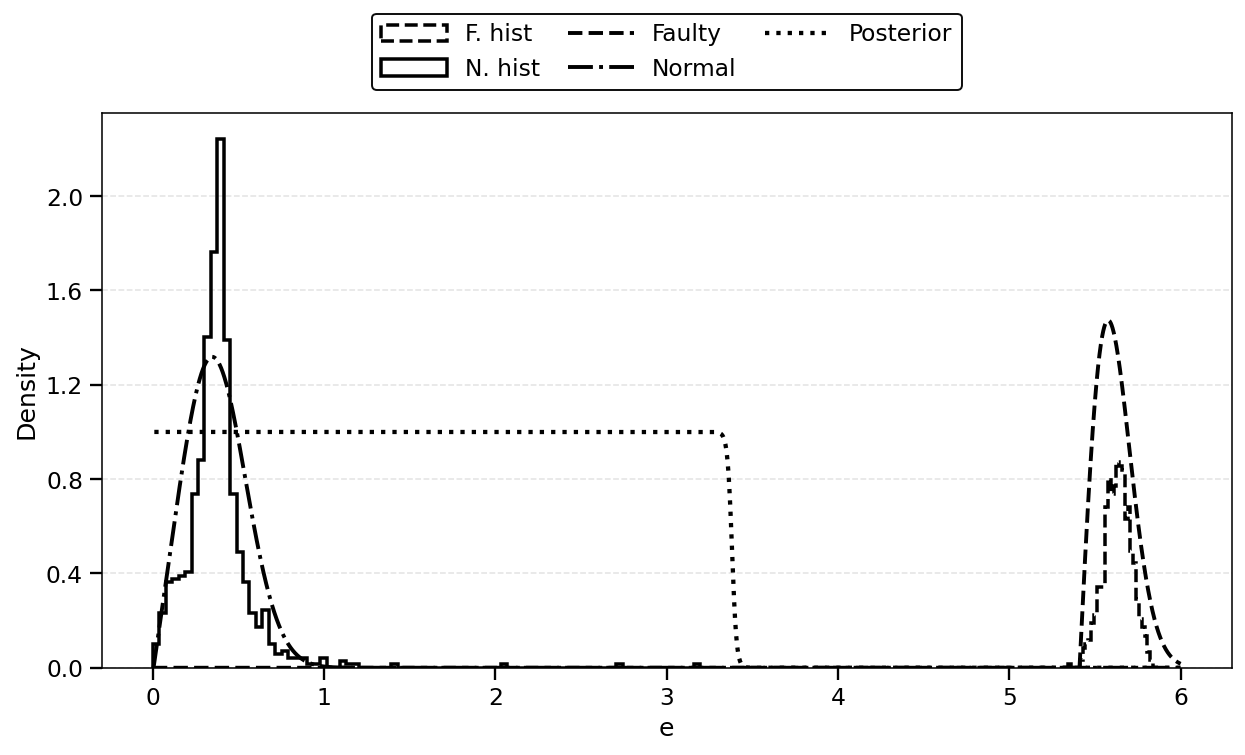}
    \caption{Results of the statistical analysis. The lines labelled as Normal and Faulty represent the estimated likelihoods of both hypotheses weighted times the corresponding priors. The line labelled as posterior corresponds to $p(N|e)$}.
    \label{fig:result_histograms}
\end{figure}

In order to estimate the marginal likelihood of the error, both conditional likelihoods have been modeled with two Rice distributions with the form

\begin{equation}\label{eq:likelihood}
    \begin{split}
    p(e|H)=&\frac{2(e-\mu)}{\alpha}\text{exp}\left(\frac{-((e-\mu_H)^2+\nu_H^2)}{\alpha_H}\right)\cdot\\
    &\cdot I_0\left(\frac{2(e-\mu_H)\nu_H}{\alpha_H}\right)
    \end{split}
\end{equation}
where parameters $\mu_H$, $\nu_H$ and $\alpha_H$ are the bias, shape and scale of the distributions, for hypothesys $H=A$ (anomaly) and $H=N$ (normal). The L-BFGS-B algorithm is used to minimize the negative log likelihood 
\begin{equation}
     NLL = - \sum_{i=0}^{N-1} \log \left( w p(e_i|N) + (1-w) p(e_i|A) \right)
\end{equation}
with respect to parameters $w, \mu_N, \nu_N, \alpha_N, \mu_N, \nu_N, \alpha_N$, where $w$ plays the role of the prior $p(N)$. In order to add robustness to the optimization, we apply a simulated annealing to the optimization. This is, we start by initializing the 7 parameters at random, and run an optimization. We repeat the optimization 100 times, and keep the optimized values that provide the 10 lowest values of the negative log likelihood. Then we generate 10 random parameter sets around these values, run the optimization with them as initial values, keep the 10 best results and iterate the process until convergence. 

Fig. \ref{fig:result_histograms} shows the result of the statistical analysis after this optimization, where both joint probability estimations $p(e,H)=p(H)p(e|H)$ are represented in a dash-dotted line for $H=N$ and in a dashed line for $H=A$. The optimized values for the parameters were $\nu_N=1.82, \mu_N=0, \alpha_N=0.17, \nu_A=0, \mu_A=2.94,\alpha_A=0.25, w=0.6$. Note here that for the case of faulty images (H=A), the shape parameter $\nu_A$ is zero, and therefore the distribution becomes a biased Rayleigh distribution. The distribution for the normal images is not biased ($\mu_N=0$). The prior of the normal images is $p(N)=w=0.6$, as expected since this is the fraction of normal images in the dataset. 

{\ Fig. \ref{fig:result_histograms} shows four images that are classified as normal, but that show errors of amplitudes 1.41, 2.03,  2.74 and 3.2.  These images contain artifacts of high amplitude, which is over 1000 times. The CAE is not able to reconstruct these amplitudes, and therefore, the error highly deviates from the error mode of the normal images. The average maximum amplitude of the images is of about 10 times lower than the amplitude of these images. The method does not distinguish such subtle anomalies, and further refinements should be applied to detect these artifacts.}

{\ Additional experiments were conducted to assess the method’s discrimination performance when discrimination is performed at the tile level during testing. This approach reduces test time by approximately a factor of three, while yielding no noticeable difference in discrimination performance.} 

\subsection{Estimation of the detection threshold}

In order to estimate the detection threshold we simply establish a minimum posterior probability for the normal images of 50\%. The MSE corresponding to this can be computed by using Eq. \eqref{eq:posterior} with Eq. \eqref{eq:likelihood} and the optimized parameters. For the above experiment this threshold is $e_t = 2.925$.  

 In order to determine the quality of the threshold, the Receiver Operating Curve (ROC) has been computed for different detection thresholds ranging from 0 to 5. This curve shows the relationship between the fraction of true positive detections and false positive detections for all the thresholds in the range. The graphic has been zoomed for a false positive rate between 0 and $10^{-2}$. The result of the threshold chosen with the algorithm above is marked with a dot in the figure, and it shows a detection rate of 1, and a false positive rate of $4.4\times 10^{-3}$. The errors correspond to 4 normal data whose estimation error is high. 

\begin{figure}
    \centering
    \includegraphics[width=\linewidth]{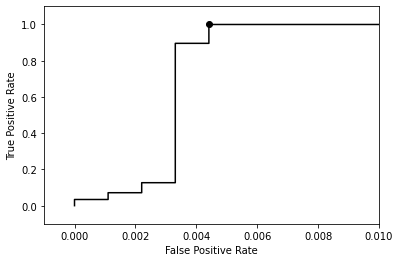}
    \caption{ROC curve for various values of the detection threshold. The dot corresponds to the automatically chosen value of the threshold, with a $100\%$ of detections and a false positive rate of $4.4\times 10^{-3}$.}
    \label{fig:enter-label}
\end{figure}

\begin{figure*}
    \centering
    \includegraphics[width=0.48\linewidth]{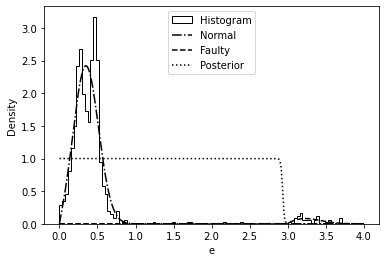} 
    \includegraphics[width=0.497\linewidth]{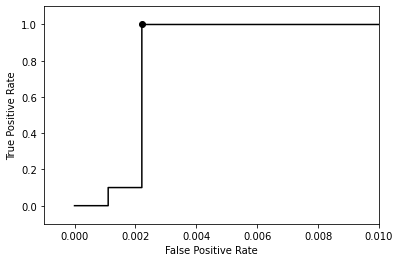} 
    \caption{Results of an automatic threshold estimation with a 98\% of normal images and a 2\% of faulty images. The dot in the left pane shows the result of the true positive rate as a function of the true positive rate for the estimated threshold. The false positive rate in this case is of  $2.2\times 10^{-3}$.}
    \label{fig:results2}
\end{figure*}

\subsection{Experiment with reduced number of anomalies}

In order to show the robustness of the method, the test is repeated with a reduced number of faulty images. The dataset used contains now the same number of normal images but only 30 faulty images chosen at random, so the proportion of normal to faulty images is now 98\% of normal  to 2\% of faulty images. The experiment is repeated to compute the new distributions and automatic threshold.

Fig. \ref{fig:results2} shows the result of the experiment. The left pane shows the histogram of the error, where the faulty images are now a small cluster. The estimated likelihoods are visually in agreement with the histograms. The posterior probability is also represented. The right pane shows the ROC with the point automatically selected, showing a false positive rate of $2.2\times 10^{-3}$ or 0.22\%. 

In this case, the optimized values for the parameters were $\nu_N=1.81, \mu_N=0, \alpha_N=0.17, \nu_A=0, \mu_A=3,\alpha_A=0.23, w=0.97$. These values are in high agreement with the previous ones, but the prior $p(N)$ represented by variable $w$ is 0.97, in agreement with the fraction of faulty data.

\section{\label{sec:conclusions}Conclusions}

We presented a methodology for anomaly detection applied to the detection of faulty diffraction patterns in single shot MUED experiments that is fully unsupervised. This is, the system is not provided with any information about what images are normal or faulty, which allows the user to train the method without the necessity of labeling the training  images as normal or faulty. This way, the method can be trained and tested with a high number of images in short time and faulty images can be removed effectively. This can lead to increased resolution in MUED by improving the signal-to-noise ratio in averaged diffraction patterns {\  and can also be employed as a stability diagnostic tool for the system by tracking the faulty detection rate. For pump-probe MUED experiments, where subtle changes are expected in the diffraction patterns, this methodology can dramatically increase the technique accuracy. This unsupervised filtering procedure could also be applied to aid users of other diffraction techniques (such as in-situ transmission electron diffraction) in which large datasets are collected that include faulty images due to instrumental instabilities. It should be noted, however, that this filtering procedure is not a substitute for stability and considers outliers as rare events.}

The method provides the user with an estimation of the probability of that the image is normal. This informs the user about the images that are difficult to classify for them to visually inspect a posterior. The detection  threshold is computed automatically using a set of validation images that do not need to be labelled either, which makes this system fully flexed and totally autonomous, freeing the user of manually adjusting any parameters of the detector. 

The methodology has been trained with only 100 images. The test images were 1521, which were used first to estimate the threshold. After this, these images were tested to detect the faulty ones, with a false positive rate that ranged between 0.2 and 0.4\%. The training time was of about 10 seconds per image, and the test time was of about one second per image. 

{\ The core anomaly detector idea is general, but our current pre-processing is tuned to crystalline patterns dominated by localized Bragg peaks. Indeed, in order to select the image areas relevant for the training, we perform a tile selection through the estimation of connected above-threshold pixels. For applications as gas-phase diffraction the same framework can apply, but the pre-processing/tiling criterion should be adapted to that purpose.} 

{\ Subtle localized defects (e.g., dark-current patches, hot pixels, partial region dropout) will increase reconstruction residuals if they are not part of the learned normal distribution. The CAE residual approach is sensitive when we aggregate errors over tiles and pixels. However, small or weak localized phenomena will be blurred if we use a global mean of the images. In these cases, it is better to use aggregations that are more robust. For example, we could use the maximum tile error, upper percentiles of tile errors, or others, while using the same probabilistic approach for the post-processing of the residuals.}

{\ Some additional work is possible to further exploit the information captured in the encoder’s latent space. Although we conducted experiments using the bottleneck output to discriminate images with a Support Vector Machine (SVM) employing a squared-exponential kernel to distinguish between normal images and anomalies, these results did not outperform the unsupervised CAE residual analysis presented in this paper. Nevertheless, it may be fruitful to use a dataset labeled by anomaly type and leverage the latent representations to train a supervised classifier for anomaly type detection.}  

\section*{Acknowledgments}
The Work was supported by the U.S. Department of Energy, Office of Science, Office of Basic Energy Sciences, Materials Science and Engineering Division, Program of Electron and Scanning Probe Microscopes, award DE-SC0021365. Funding was available through the Department of Energy’s Established Program to Stimulate Competitive Research (EPSCoR) State-National Laboratory Partnerships program in the Office of Basic Energy Sciences. This work was also partially supported by the U.S. National Science Foundation under Award PHY-1549132, the Center for Bright Beams. This research used resources of the Argonne Leadership Computing Facility, which is a DOE Office of Science User Facility supported under Contract DE-AC02-06CH11357. This research used resources of the UNM Center for Advanced Research Computing, supported in part by the National Science Foundation, for providing the high-performance computing, large-scale storage, and visualization resources used in this work. This research used resources of Brookhaven National Laboratory’s Accelerator Test Facility, an Office of Science User Facility. 
Manel Martínez-Ramón has been partly supported by the King Felipe VI Endowed Chair of the University of New Mexico.

\bibliographystyle{unsrtnat}
\bibliography{references}

@inproceedings{wang2003femto,
  title={Femto-seconds electron beam diffraction using photocathode RF gun},
  author={Wang, XJ and Wu, Z and Ihee, Hyotcherl Harry},
  booktitle={Proceedings of the 2003 particle accelerator conference},
  volume={1},
  pages={420--422},
  year={2003},
  organization={IEEE}
}

@article{zhu2015femtosecond,
  title={Femtosecond time-resolved MeV electron diffraction},
  author={Zhu, Pengfei and Zhu, Y and Hidaka, Y and Wu, L and Cao, J and Berger, H and Geck, J and Kraus, R and Pjerov, S and Shen, Y and others},
  journal={New Journal of Physics},
  volume={17},
  number={6},
  pages={063004},
  year={2015},
  publisher={IOP Publishing}
}

@article{kealhofer2015signal,
  title={Signal-to-noise in femtosecond electron diffraction},
  author={Kealhofer, Catherine and Lahme, Stefan and Urban, Theresa and Baum, Peter},
  journal={Ultramicroscopy},
  volume={159},
  pages={19--25},
  year={2015},
  publisher={Elsevier}
}

@article{mao2016image,
  title={Image restoration using very deep convolutional encoder-decoder networks with symmetric skip connections},
  author={Mao, Xiaojiao and Shen, Chunhua and Yang, Yu-Bin},
  journal={Advances in neural information processing systems},
  volume={29},
  year={2016}
}

@article{gu2018recent,
  title={Recent advances in convolutional neural networks},
  author={Gu, Jiuxiang and Wang, Zhenhua and Kuen, Jason and Ma, Lianyang and Shahroudy, Amir and Shuai, Bing and Liu, Ting and Wang, Xingxing and Wang, Gang and Cai, Jianfei and others},
  journal={Pattern recognition},
  volume={77},
  pages={354--377},
  year={2018},
  publisher={Elsevier}
}

@article{fukushima1982neocognitron,
  title={Neocognitron: A new algorithm for pattern recognition tolerant of deformations and shifts in position},
  author={Fukushima, Kunihiko and Miyake, Sei},
  journal={Pattern recognition},
  volume={15},
  number={6},
  pages={455--469},
  year={1982},
  publisher={Elsevier}
}

@article{lecun1998gradient,
  title={Gradient-based learning applied to document recognition},
  author={LeCun, Yann and Bottou, L{\'e}on and Bengio, Yoshua and Haffner, Patrick},
  journal={Proceedings of the IEEE},
  volume={86},
  number={11},
  pages={2278--2324},
  year={1998},
  publisher={Ieee}
}

@inproceedings{nair2010rectified,
  title={Rectified linear units improve restricted boltzmann machines},
  author={Nair, Vinod and Hinton, Geoffrey E},
  booktitle={Proceedings of the 27th international conference on machine learning (ICML-10)},
  pages={807--814},
  year={2010}
}

@article{dumoulin2016guide,
  title={A guide to convolution arithmetic for deep learning},
  author={Dumoulin, Vincent and Visin, Francesco},
  journal={arXiv preprint arXiv:1603.07285},
  year={2016}
}

@article{hwang2009survey,
  title={A survey of fault detection, isolation, and reconfiguration methods},
  author={Hwang, Inseok and Kim, Sungwan and Kim, Youdan and Seah, Chze Eng},
  journal={IEEE transactions on control systems technology},
  volume={18},
  number={3},
  pages={636--653},
  year={2009},
  publisher={IEEE}
}

@article{cao2019generalization,
  title={Generalization bounds of stochastic gradient descent for wide and deep neural networks},
  author={Cao, Yuan and Gu, Quanquan},
  journal={Advances in neural information processing systems},
  volume={32},
  year={2019}
}

@book{murphy2012machine,
  title={Machine learning: a probabilistic perspective},
  author={Murphy, Kevin P},
  year={2012},
  publisher={MIT press}
}

@book{fletcher2000practical,
  title={Practical methods of optimization},
  author={Fletcher, Roger},
  year={2000},
  publisher={John Wiley \& Sons}
}

@book{shawe2004kernel,
  title={Kernel methods for pattern analysis},
  author={Shawe-Taylor, John and Cristianini, Nello and others},
  year={2004},
  publisher={Cambridge university press}
}

@book{vapnik2013nature,
  title={The nature of statistical learning theory},
  author={Vapnik, Vladimir},
  year={2013},
  publisher={Springer science \& business media}
}

@article{goldstein2012histogram,
  title={Histogram-based outlier score (hbos): A fast unsupervised anomaly detection algorithm},
  author={Goldstein, Markus and Dengel, Andreas},
  journal={KI-2012: Poster and Demo Track},
  pages={59--63},
  year={2012},
  publisher={Citeseer}
}

@inproceedings{breunig2000lof,
  title={LOF: identifying density-based local outliers},
  author={Breunig, Markus M and Kriegel, Hans-Peter and Ng, Raymond T and Sander, J{\"o}rg},
  booktitle={Proceedings of the 2000 ACM SIGMOD international conference on Management of data},
  pages={93--104},
  year={2000}
}

@inproceedings{ramaswamy2000efficient,
  title={Efficient algorithms for mining outliers from large data sets},
  author={Ramaswamy, Sridhar and Rastogi, Rajeev and Shim, Kyuseok},
  booktitle={Proceedings of the 2000 ACM SIGMOD international conference on Management of data},
  pages={427--438},
  year={2000}
}

@inproceedings{angiulli2002fast,
  title={Fast outlier detection in high dimensional spaces},
  author={Angiulli, Fabrizio and Pizzuti, Clara},
  booktitle={European conference on principles of data mining and knowledge discovery},
  pages={15--27},
  year={2002},
  organization={Springer}
}

@inproceedings{lazarevic2005feature,
  title={Feature bagging for outlier detection},
  author={Lazarevic, Aleksandar and Kumar, Vipin},
  booktitle={Proceedings of the eleventh ACM SIGKDD international conference on Knowledge discovery in data mining},
  pages={157--166},
  year={2005}
}

@article{aggarwal2015theoretical,
  title={Theoretical foundations and algorithms for outlier ensembles},
  author={Aggarwal, Charu C and Sathe, Saket},
  journal={Acm sigkdd explorations newsletter},
  volume={17},
  number={1},
  pages={24--47},
  year={2015},
  publisher={ACM New York, NY, USA}
}

@inproceedings{sakurada2014anomaly,
  title={Anomaly detection using autoencoders with nonlinear dimensionality reduction},
  author={Sakurada, Mayu and Yairi, Takehisa},
  booktitle={Proceedings of the MLSDA 2014 2nd workshop on machine learning for sensory data analysis},
  pages={4--11},
  year={2014}
}

@inproceedings{chen2017outlier,
  title={Outlier detection with autoencoder ensembles},
  author={Chen, Jinghui and Sathe, Saket and Aggarwal, Charu and Turaga, Deepak},
  booktitle={Proceedings of the 2017 SIAM international conference on data mining},
  pages={90--98},
  year={2017},
  organization={SIAM}
}

@article{hinton2006reducing,
  title={Reducing the dimensionality of data with neural networks},
  author={Hinton, Geoffrey E and Salakhutdinov, Ruslan R},
  journal={science},
  volume={313},
  number={5786},
  pages={504--507},
  year={2006},
  publisher={American Association for the Advancement of Science}
}

@article{hardin2004outlier,
  title={Outlier detection in the multiple cluster setting using the minimum covariance determinant estimator},
  author={Hardin, Johanna and Rocke, David M},
  journal={Computational Statistics \& Data Analysis},
  volume={44},
  number={4},
  pages={625--638},
  year={2004},
  publisher={Elsevier}
}

@techreport{shyu2003novel,
  title={A novel anomaly detection scheme based on principal component classifier},
  author={Shyu, Mei-Ling and Chen, Shu-Ching and Sarinnapakorn, Kanoksri and Chang, LiWu},
  year={2003},
  institution={Dept. of Electrical and Computer Engineering, Miami University,  Coral Gables, FL }
}

@article{scholkopf2001estimating,
  title={Estimating the support of a high-dimensional distribution},
  author={Sch{\"o}lkopf, Bernhard and Platt, John C and Shawe-Taylor, John and Smola, Alex J and Williamson, Robert C},
  journal={Neural computation},
  volume={13},
  number={7},
  pages={1443--1471},
  year={2001},
  publisher={MIT Press One Rogers Street, Cambridge, MA 02142-1209, USA journals-info~…}
}

@article{tax2004support,
  title={Support vector data description},
  author={Tax, David MJ and Duin, Robert PW},
  journal={Machine learning},
  volume={54},
  number={1},
  pages={45--66},
  year={2004},
  publisher={Springer}
}

@article{goodfellow2020generative,
  title={Generative adversarial networks},
  author={Goodfellow, Ian and Pouget-Abadie, Jean and Mirza, Mehdi and Xu, Bing and Warde-Farley, David and Ozair, Sherjil and Courville, Aaron and Bengio, Yoshua},
  journal={Communications of the ACM},
  volume={63},
  number={11},
  pages={139--144},
  year={2020},
  publisher={ACM New York, NY, USA}
}

@article{zamzam2019data,
	title={Data-driven learning-based optimization for distribution system state estimation},
	author={Zamzam, Ahmed S and Fu, Xiao and Sidiropoulos, Nicholas D},
	journal={IEEE Transactions on Power Systems},
	volume={34},
	number={6},
	pages={4796--4805},
	year={2019},
	publisher={IEEE}
}

@article{liu2019generative,
  title={Generative adversarial active learning for unsupervised outlier detection},
  author={Liu, Yezheng and Li, Zhe and Zhou, Chong and Jiang, Yuanchun and Sun, Jianshan and Wang, Meng and He, Xiangnan},
  journal={IEEE Transactions on Knowledge and Data Engineering},
  volume={32},
  number={8},
  pages={1517--1528},
  year={2019},
  publisher={IEEE}
}

@article{weathersby2015mega,
  title={Mega-electron-volt ultrafast electron diffraction at {SLAC National Accelerator Laboratory}},
  author={Weathersby, SP and Brown, G and Centurion, Martin and Chase, TF and Coffee, Ryan and Corbett, Jeff and Eichner, JP and Frisch, JC and Fry, AR and G{\"u}hr, M and others},
  journal={Review of Scientific Instruments},
  volume={86},
  number={7},
  year={2015},
  publisher={AIP Publishing}
}

@article{qi2020breaking,
  title={Breaking 50 femtosecond resolution barrier in {MeV} ultrafast electron diffraction with a double bend achromat compressor},
  author={Qi, Fengfeng and Ma, Zhuoran and Zhao, Lingrong and Cheng, Yun and Jiang, Wenxiang and Lu, Chao and Jiang, Tao and Qian, Dong and Wang, Zhe and Zhang, Wentao and others},
  journal={Physical review letters},
  volume={124},
  number={13},
  pages={134803},
  year={2020},
  publisher={APS}
}

@article{baldini2023spontaneous,
  title={The spontaneous symmetry breaking in {Ta\textsubscript{2}NiSe\textsubscript{5}} is structural in nature},
  author={Baldini, Edoardo and Zong, Alfred and Choi, Dongsung and Lee, Changmin and Michael, Marios H and Windgaetter, Lukas and Mazin, Igor I and Latini, Simone and Azoury, Doron and Lv, Baiqing and others},
  journal={Proceedings of the National Academy of Sciences},
  volume={120},
  number={17},
  pages={e2221688120},
  year={2023},
  publisher={National Academy of Sciences}
}

@article{chen2025structural,
  title={Structural Contribution to Light-Induced Gap Suppression in {Ta\textsubscript{2}NiSe\textsubscript{5}}},
  author={Chen, Zijing and Xu, Chenhang and Xie, Chendi and Tang, Weichen and Liu, Qiaomei and Wu, Dong and Xu, Qing and Jiang, Tao and Zhu, Pengfei and Zou, Xiao and others},
  journal={Physical Review Letters},
  volume={135},
  number={9},
  pages={096901},
  year={2025},
  publisher={APS}
}

@article{kingma2014adam,
  title={Adam: A method for stochastic optimization},
  author={Kingma, Diederik P},
  journal={arXiv preprint arXiv:1412.6980},
  year={2014}
}

\end{document}